\documentclass[conference]{IEEEtran}
\IEEEoverridecommandlockouts
\usepackage{cite}
\usepackage{hyperref}
\usepackage{amsmath,amssymb,amsfonts}
\usepackage{mathtools}

\usepackage{algorithmic}
\usepackage{graphicx}
\usepackage{textcomp}
\usepackage{xcolor}
\usepackage{svg}
\usepackage[]{subcaption}
\usepackage{xcolor}
\usepackage{epsfig} 
\def\BibTeX{{\rm B\kern-.05em{\sc i\kern-.025em b}\kern-.08em
    T\kern-.1667em\lower.7ex\hbox{E}\kern-.125emX}}
\newcommand{\Mod}[1]{\ \mathrm{mod}\ #1}    

\begin{document}

\title{Bandwidth-Adaptive Feature Sharing for Cooperative LIDAR Object Detection
\thanks{This research was supported by the National Science Foundation under CAREER Grant CNS-1664968.}

}
\author{\IEEEauthorblockN{Ehsan Emad Marvasti\IEEEauthorrefmark{1}, Arash Raftari\IEEEauthorrefmark{1}, Amir Emad Marvasti\IEEEauthorrefmark{1}, Yaser P. Fallah\IEEEauthorrefmark{1}}
\IEEEauthorblockA{\IEEEauthorrefmark{1}Department of Electrical and Computer Engineering, Department of Computer Science\\
University of Central Florida, Orlando, Florida, USA\\
\{e\_emad, raftari, a\_emad\}@knights.ucf.edu} yaser.fallah@ucf.edu

}

\maketitle

\begin{abstract}
Situational awareness as a necessity in the connected and autonomous vehicles (CAV) domain is the subject of a significant number of researches in recent years. The driver’s safety is directly dependent on the robustness, reliability, and scalability of such systems. Cooperative mechanisms have provided a solution to improve situational awareness by utilizing high speed wireless vehicular networks. These mechanisms mitigate problems such as occlusion and sensor range limitation. However, the network capacity is a factor determining the maximum amount of information being shared among cooperative entities. The notion of feature sharing, proposed in our previous work, aims to address these challenges by maintaining a balance between computation and communication load. In this work, we propose a mechanism to add flexibility in adapting to communication channel capacity and a novel decentralized shared data alignment method to further improve cooperative object detection performance. The performance of the proposed framework is verified through experiments on Volony dataset. The results confirm that our proposed framework outperforms our previous cooperative object detection method (FS-COD) in terms of average precision.
\end{abstract}


\begin{figure*}[!t]
\centering
 \includegraphics[width=1\textwidth,trim={0mm 0mm 0mm 0mm},clip]{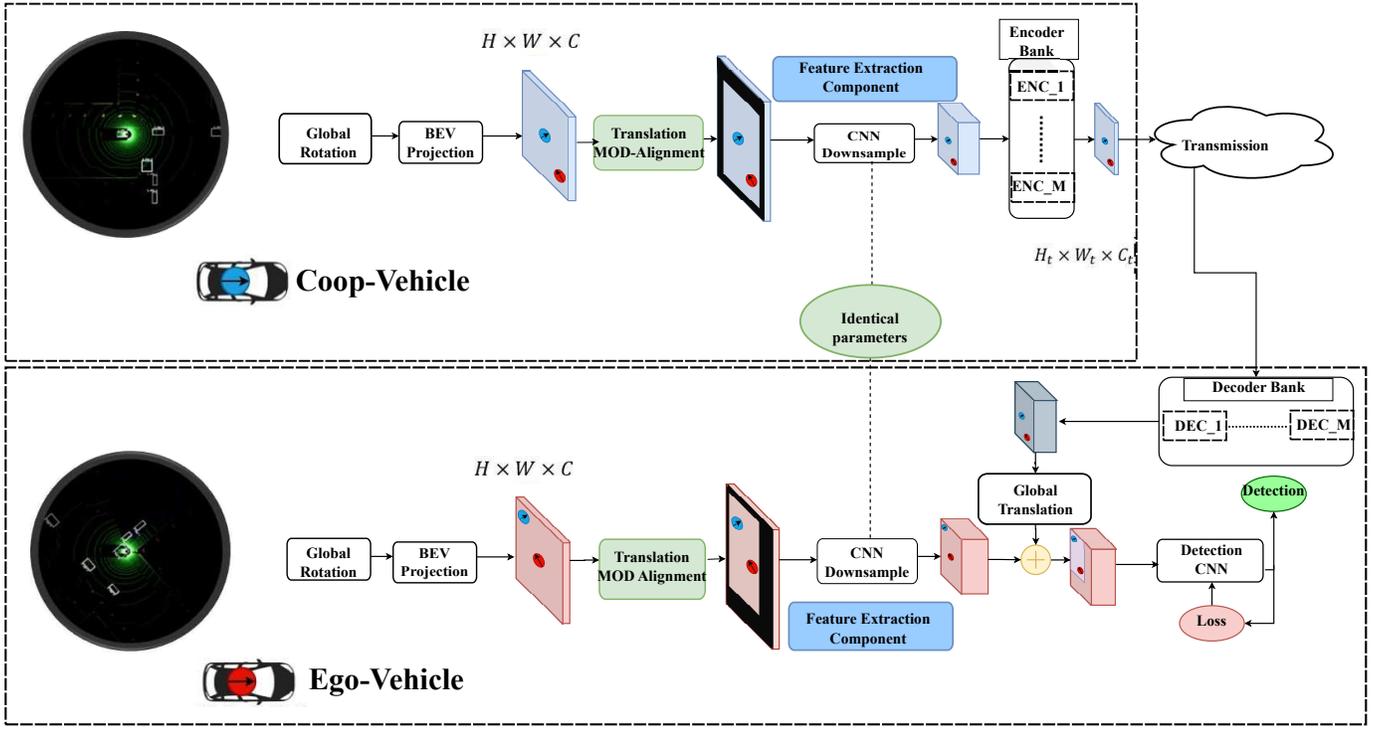}
 \caption{The overview of AFS-COD architecture}
 \label{fig:overall}
 \end{figure*}
\section{Introduction}
The recent advancement in computational systems has led to tremendous progress in the performance and functionality of Advanced Driving Assistance Systems (ADAS).
Since the reliability of such systems is dependent on the robustness of on-board sensory and perception units, there have been constant efforts to improve the quality of sensory devices and perception inference systems.
Object detectors as one of the main components of perception inference systems have been vastly studied.
Many high performing and relatively fast Convolutional Neural Network (CNN) based object detection methods have been introduced in recent years \cite{Redmon_2016_CVPR,li2016vehicle,feng2018towards,beltran2018birdnet,10.1007/978-3-319-46448-0_2}. Despite of these advancements, the performance of these methods is strictly dependent on the quality of input data provided by on-board sensory devices. Therefore, their performance severely deteriorates in the detection of non-line-of-sight or occluded objects if they are utilized in single-vehicle object detection setup. In order to alleviate the aforementioned challenges, Vehicle to Everything (V2X) communication technology can be considered as a strong and appealing candidate.

 V2X networks offer the possibility of cooperation amongst road participants, equipped with on-board perception systems, to expand range of sensing, alleviate the non-line-of-sight restrictions and possibly achieve synergy in sensing performance.  These advantages can lead to improvement in road participants' situational awareness through designing distributed sensing/perception systems, i.e., cooperative perception methods. However, scalability issues caused by limited communication bandwidth should be considered for designing any cooperative perception method.
 
Generally, cooperative perception methods can be distinguished by the type of information shared amongst participants.
The shared information can be categorized to raw, partially processed, and fully processed.
While methods based on raw and fully processed information sharing have been vastly studied; the approach of sharing partially processed information is relatively new. %

Concepts such as Cooperative Sensing Message (CSM), Collective Perception message (CPM), and Environmental Perception Message (EPM), introduced and investigated in \cite{CSM},\cite{CPM} and \cite{7835930}, are attempts for achieving cooperative perception by disseminating fully processed information (object detection results).
Although these methods are different in proposed packet formation, they provide a framework for transmission and aggregation of fully processed information.

In addition, there are several efforts, in particular from collaborative sensing perspective, approaching cooperative perception by sharing raw information. The multimodal cooperative perception to provide a far sight satellite view of the environment to the driver proposed in \cite{6866903} or Augmented vehicular reality method to broadens cooperative vehicles' visual horizon presented in \cite{10.1145/3210240.3210319} are examples of this approach.

While, in general, cooperative perception techniques by sharing raw information have shown more promising performance; these techniques require significantly larger communication bandwidth. On the other hand, although techniques based on sharing fully processed data require much less communication capacity, detection of partially observed objects and the lack of consensus between cooperative entities are challenging to be addressed by these techniques. 

We present the general concept of Feature Integration for Cooperative Sensing (FICS) in this paper. In FICS, the main data units that are exchanged between cooperating nodes are the features that are extracted from intermediate layers of a multi layer neural network (such as a CNN). In the FICS framework, the essential tasks consist of extracting features, sharing the features with other cooperative nodes, and aligning and integrating the shared features.
A realization of FICS for object detection using LIDAR is our earlier work proposed in \cite{marvasti2020cooperative}: Feature sharing cooperative object detection (FS-COD). Another method utilizing sharing of partially processed data (features) in LIDAR object detection is feature based cooperative perception (F-COOPER)\cite{10.1145/3318216.3363300}.
These methods attempt to find a middle ground between rich information content of sharing raw data and low communication bandwidth requirement of sharing fully processed data.

Although the aforementioned FS-COD method is capable of reducing the required communication bandwidth, in order to aggregate the shared feature-maps and perform object detection, this technique requires the cooperative entities to produce and share feature-maps with a specific pre-defined size. The required communication bandwidth in such design is proportional to this size and consequently, is dependent on the neural network architecture, i.e., feature extractor layer. In this work, we propose a new framework, called Adaptive Feature Sharing for Cooperative Object Detection (AFS-COD) to disentangle the bandwidth limitation from the architectural design and enable cooperative vehicles to share feature-maps with variable sizes. Additionally, we have identified a shared data alignment error arising by down-sampling and propose a novel two-step decentralized translation alignment method to address this issue and enhance cooperative object detection performance. AFS-COD is a realization of FICS for object detection using LIDAR based data.

The rest of this paper is organized as follows. In section \ref{section:Variable Feature Sharing Cooperative Object Detection (AFS-COD)}, the overall framework is introduced and its main building blocks including the decentralized two-step alignment procedure are elaborated. In section \ref{Section:Experiments and Results}, the dataset and the baseline methods are concisely discussed and the results of our proposed framework are compared with baseline methods. Finally, section \ref{Section:Concluding Remarks and Future work} concludes this work. 

\section{Adaptive Feature Sharing Cooperative Object Detection (AFS-COD)}
\label{section:Variable Feature Sharing Cooperative Object Detection (AFS-COD)}
The promising result of FS-COD in improving situational awareness motivated us to further investigate the capabilities provided by feature sharing concept.
FS-COD, as a preliminary baseline architecture, provides insight on different necessary components to efficiently train, align, and aggregate the information gathered from LIDAR devices by cooperative vehicles. 
However, many components in FS-COD can be subjected to further investigation and improvement.
These components mainly consist of input representation, shared-data alignment procedure, feature extraction, feature fusion, training procedure, and loss function design. 

From these components, the focus of this paper is on feature transmission and fusion components and shared-data alignment procedure. The motivation behind the first modification is to provide an architecture capable of adapting to various network bandwidth limitations by enabling the cooperative vehicles to transmit and aggregate feature-maps with different sizes. The rationale for the second improvement is to enhance cooperative object detection performance by introducing Translation MOD-Alignment procedure. This procedure aims to eliminate inherent localization error caused by down-sampling of input data in feature extraction layers without the need of prior knowledge on transmitter's position.
\begin{figure*}[!t]%
\centering
\subcaptionbox[]{\label{fig:MA}}{%
\includegraphics[width=0.40\textwidth,trim={0mm 0mm 0mm 0mm},clip]{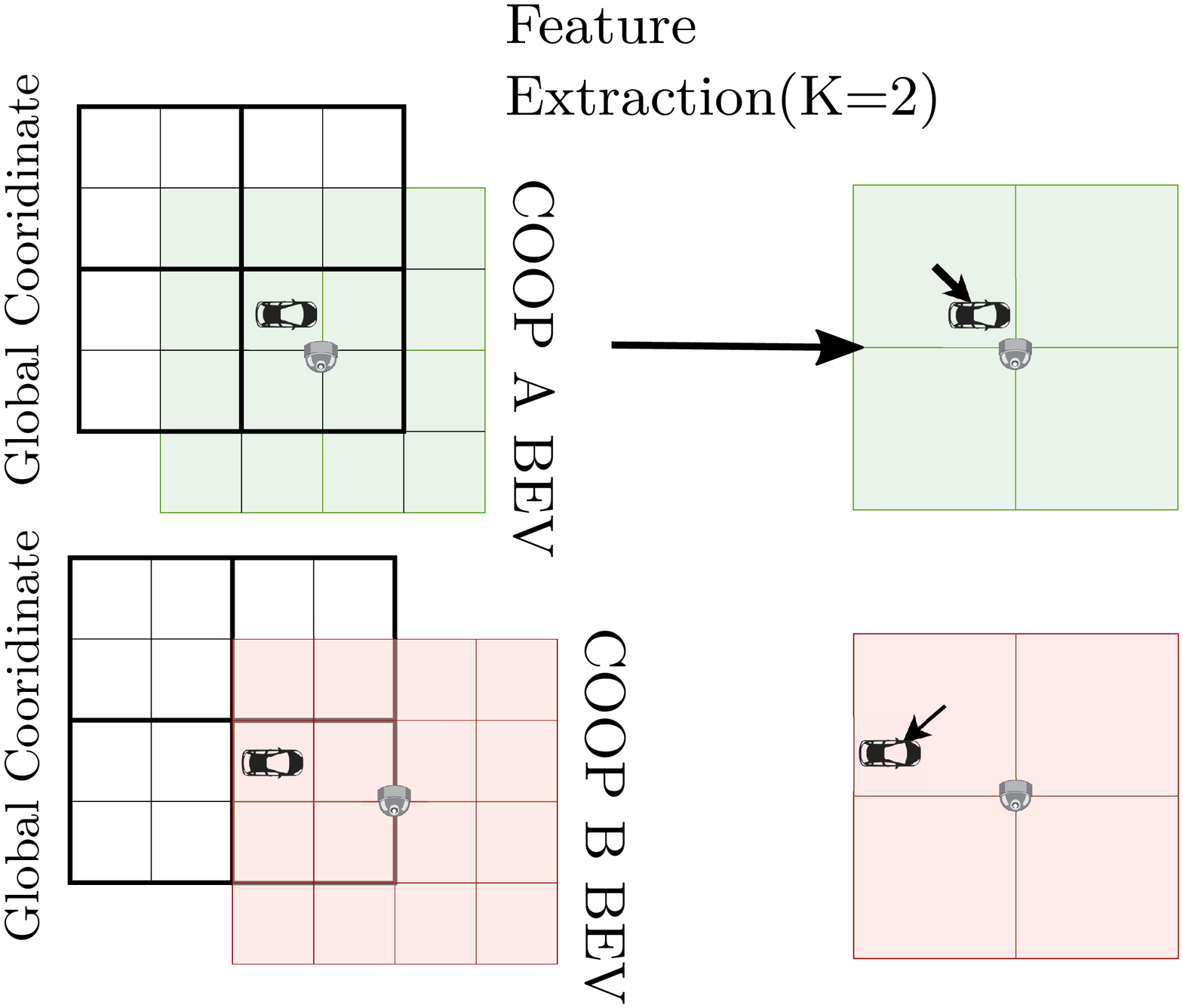}}%
\subcaptionbox[]{\label{fig:TMA}}{%
\includegraphics[width=0.59\textwidth,trim={5mm 25mm 10mm 15mm},clip]{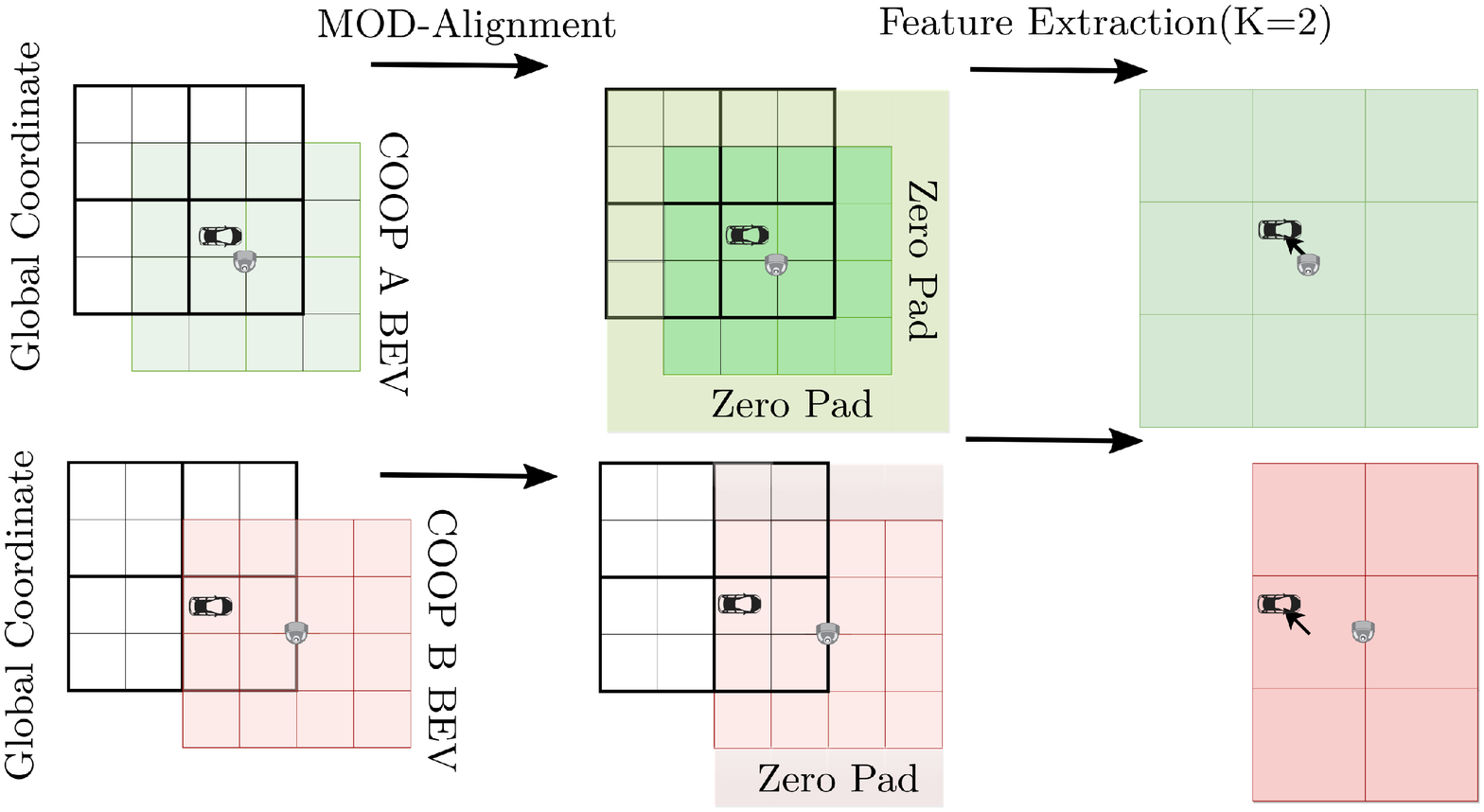}}%


\caption{ (a) An illustration of the effect of down-sampling on alignment. The vectors pointing to the target within the corresponding fixels are contradictory. This shows that a simple shift in the observation image would change the produced feature maps. K is the down-sampling rate (b) An illustration of how the Translation MOD-Alignment procedure resolves the information mismatch caused down-sampling. The input image is padded with zero values to ensure each produced fixel represents a specific range of pixels in global coordinate system. The vectors pointing to the target object within both corresponding fixels are identical.}
\label{fig:foobar}
\end{figure*}

It is worth mentioning that the problem motivating the introduction of Translation MOD-Alignment is not unique in FS-COD but rather in any method of information fusion where grids of down-sampled observations are aggregated. Therefore, such a procedure may be applicable to other cooperative information fusion applications.

\subsection{AFS-COD Architecture}

In this section, an overview of AFS-COD method along with a brief description of its components is presented.
In the proposed setup, we assume that each cooperative vehicle is equipped with LIDAR as a sensory unit and GPS device for reading their position information.
Therefore, the observations (input data) are the point-clouds generated from the sensory unit. Additionally, cooperative vehicles share the partially processed information (features) along the metadata containing their position information. 
In general, any CNN based object detection method is distinguished by optimization cost functions, network architecture, and input representation.
Similar to FS-COD, in the proposed AFS-COD framework, the object detection component was designed by adapting the loss function presented in \cite{Redmon_2016_CVPR} and bird-eye view (BEV) projection of point-clouds is chosen as data representation due to its merits in our specific vehicular application \cite{Yang_2018_CVPR,kim2019enhanced}.
Fig. \ref{fig:overall} demonstrates the overview of AFS-COD framework.

\subsubsection{Global Rotation Alignment}

The main purpose of feature sharing concept is to enable participating entities to combine corresponding features extracted separately by cooperative vehicles. In order to achieve this goal, the extracted feature-maps should be aligned before being combined. Similarly to FS-COD we follow the same procedure and align the point clouds with respect to a predefined global coordinate system using rotation transformation.

\subsubsection{BEV Projection}
As it was discussed, in our proposed setup, the 2D images, obtained by BEV projection of LIDAR point-clouds, are fed to neural networks as input data. After the generated point-clouds are globally aligned with respect to rotation, BEV projector unit projects the aligned LIDAR point-clouds onto 2D image plane. In this work, BEV image has 3 channels and each channel represents the density of reflected point clouds captured within the following height bins: $[-1m,1m]$, $[1m,3m]$ and $[3m,5m]$.

\begin{table}[t]
\caption{The architecture of Proposed networks}
\label{archtab}
\begin{center}
\begin{tabular}{ |c | c|| c|}

\hline
 \textbf{FS-COD-NMA}&\textbf{FS-COD-TMA}&\textbf{AFS-COD} \\
\hline
\multicolumn{3}{|c|}{Input 832x832x3 } \\ \hline \hline
\multicolumn{3}{|c|}{\textbf{Feature Extraction Component}} \\ \hline
\multicolumn{3}{|c|}{3x3x24 Convolution Batch-Norm Leaky ReLU(0.1)} \\ \hline
\multicolumn{3}{|c|}{Maxpool/2} \\ \hline
\multicolumn{3}{|c|}{3x3x48 Convolution Batch-Norm Leaky ReLU(0.1)}\\ \hline
\multicolumn{3}{|c|}{Maxpool/2} \\ \hline
\multicolumn{3}{|c|}{3x3x64 Convolution Batch-Norm Leaky ReLU(0.1)} \\ \hline
\multicolumn{3}{|c|}{3x3x32 Convolution Batch-Norm Leaky ReLU(0.1)} \\ \hline
\multicolumn{3}{|c|}{3x3x64 Convolution Batch-Norm Leaky ReLU(0.1)} \\ \hline
\multicolumn{3}{|c|}{Maxpool/2} \\ \hline
\multicolumn{3}{|c|}{3x3x128 Convolution Batch-Norm Leaky ReLU(0.1)} \\ \hline
\multicolumn{3}{|c|}{3x3x64 Convolution Batch-Norm Leaky ReLU(0.1)} \\ \hline
\multicolumn{3}{|c|}{3x3x128 Convolution Batch-Norm Leaky ReLU(0.1)} \\ \hline
\multicolumn{3}{|c|}{Maxpool/2} \\ \hline
\multicolumn{3}{|c|}{3x3x128 Convolution Batch-Norm Leaky ReLU(0.1)} \\ \hline
\multicolumn{2}{|c|}{3x3x$C_t$}&\multicolumn{1}{||c|}{3x3x128} \\\hline
\multicolumn{2}{|c|}{-}&\multicolumn{1}{||c|}{\textbf{Encoder}} \\ \hline
\multicolumn{2}{|c|}{-}&\multicolumn{1}{||c|}{1x1x128 Conv BN Leaky} \\ \hline
\multicolumn{2}{|c|}{-}&\multicolumn{1}{||c|}{1x1x128 Conv BN Leaky} \\ \hline
\multicolumn{2}{|c|}{-}&\multicolumn{1}{||c|}{1x1x$C_t$ Conv BN Leaky} \\ \hline
\multicolumn{2}{|c|}{-}&\multicolumn{1}{||c|}{\textbf{Decoder}} \\ \hline
\multicolumn{2}{|c|}{-}&\multicolumn{1}{||c|}{1x1x128 Conv BN Leaky} \\ \hline
\multicolumn{2}{|c|}{-}&\multicolumn{1}{||c|}{1x1x128 Conv BN Leaky} \\ \hline
\multicolumn{2}{|c|}{-}&\multicolumn{1}{||c|}{1x1x128 Conv BN Leaky} \\ \hline
\multicolumn{3}{|c|}{\textbf{Object Detection Component}} \\ \hline \hline
\multicolumn{3}{|c|}{1x1x128 Convolution Batch-Norm Leaky ReLU(0.1)}\\ \hline
\multicolumn{3}{|c|}{3x3x256 Convolution Batch-Norm Leaky ReLU(0.1)}\\ \hline
\multicolumn{3}{|c|}{1x1x512 Convolution Batch-Norm Leaky ReLU(0.1)}\\ \hline
\multicolumn{3}{|c|}{1x1x1024 Convolution Batch-Norm Leaky ReLU(0.1)}\\ \hline
\multicolumn{3}{|c|}{3x3x2048 Convolution Batch-Norm Leaky ReLU(0.1)}\\ \hline
\multicolumn{3}{|c|}{1x1x1024 Convolution Batch-Norm Leaky ReLU(0.1)} \\ \hline
\multicolumn{3}{|c|}{1x1x2048 Convolution Batch-Norm Leaky ReLU(0.1)}\\ \hline
\multicolumn{3}{|c|}{3x3x1024 Convolution Batch-Norm Leaky ReLU(0.1)} \\ \hline
\multicolumn{3}{|c|}{1x1x20 Convolution}\\ \hline
\multicolumn{3}{|c||}{Output 52x52x20}\\ \hline
\end{tabular}
\end{center}

\end{table}
\subsubsection{Translation MOD-Alignment}
Although Global Rotation and Translation alignments, originally utilized in \cite{marvasti2020cooperative}, can enable feature-maps produced by cooperative entities to be accumulated, Global Translation Alignment does not consider inconsistencies caused by down-sampling the input image. In this section, the problem in translation alignment arising from down-sampling followed by the proposed Translation MOD-Alignment method is described. For the rest of the paper, pixels produced in feature-maps are referred to as “fixels”.

Having down-sampling layers in CNN architecture a fixel in a given feature-map represent a set of pixels in input BEV image. Therefore, each fixel represents an area in observers environment.
The purpose of alignment components is to ensure that the corresponding fixels of two feature-maps are representing an identical area in the environment. 
If corresponding fixels acquired from both cooperative parties do not correspond to the same area in global coordinate, the information of such fixels are not fully compatible. Hence, the error in alignment affects the localization performance, if the feature-maps are accumulated by element-wise summation. 
Fig. \ref{fig:MA} provides an insight on how down-sampling leads to alignment error. In this figure, a target is observed by two sensors producing green and red feature-maps.
The vectors within the top left fixels in both feature-maps represents information regarding a localization vector correctly pointing to target position. 
However, if these feature-maps are aggregated and fed to an object detector the result will be erroneous due to the contradictory localization information embedded in these fixels.

If the information about the position of cooperative vehicles is available, we can align cooperative vehicles' BEV images by global rotation, translation, and padding procedures prior to feature extraction. 
However, since feature maps are extracted prior to transmission and reception, accessing the cooperative vehicles' positions is a non-realistic assumption.
Moreover, the feature-map, extracted out of BEV images aligned with such a strategy, is intended to be shared with a specific cooperative vehicle. Using this feature-map by other cooperative vehicles will lead to error in localization as explained.
We propose Translation MOD-Alignment procedure as a solution for the misalignment issue of shared feature-maps which does not require any prior knowledge on the position of cooperative vehicles. Additionally, feature maps extracted by MOD-aligned BEV images can be broadcasted and used by all cooperative entities without introducing down-sampling misalignment error.

The global pixel-wise coordinate of BEV pixels can be calculated from the observer's global coordinate position, BEV image range in meters, and resolution in pixels.
Consider the input BEV image represents global pixel-wise coordinates within the range of $[x_{0},x_{1}]\times [y_{0},y_{1}]$.
This range can be rewritten as $[\hat{x}_{0}K+\alpha_0,\hat{x}_{1}K+\alpha_1]\times [\hat{y}_{0}K+\beta_0,\hat{y}_{1}K+\beta_1]$, where $0 \leq \alpha_i,\beta_i < K;  i \in\{0, 1\} $, $\hat{x}_1>\hat{x}_0$, $\hat{y}_1>\hat{y}_0$ and $K$ is the down-sampling rate. 
Therefore the aforementioned range represent the range $[\hat{x}_{0},\hat{x}_{1}]\times [\hat{y}_{0},\hat{y}_{1}]$ in global fixel coordinates.

In this method, we pad the input BEV image in order for each produced fixel to represent a predetermined area of global coordinate.
We pad the image along both axes with zero values in order for the resulting BEV image to represent $[\hat{x}_{0}K,\hat{x}_{1}K)\times [\hat{y}_{0}K,\hat{y}_{1}K)$.
By padding the BEV image, we ensure that each produced fixel represents a unified area in the environment. The BEV image padding parameters can be calculated using the following equations.
    
\begin{equation}
\centering    
(p_l,p_t)=(\alpha_0,\beta_0) \equiv(x_0,y_0) \Mod{K} 
\end{equation}
  \begin{equation}
  \centering
(p_r,p_b)=(K-\alpha_1,K-\beta_1) \equiv (-x_1,-y_1)\Mod{K}
\end{equation}
$p_l$, $p_r$, $p_t$ and $p_b$ are left, right, top and bottom padding parameters respectively. Alternatively, the input image can be shifted to right and bottom by $p_l$ and $p_t$ if the user desires to transmit fixed-size feature map in terms of height and width. However, this will lead to loss of information on the right and bottom sides of the image. Fig. \ref{fig:TMA} illustrates an example where MOD-Alignment resolves the down-sampling issue. 

The Translation MOD-Alignment is agnostic to the position of receiver-vehicle. This property makes this method of alignment a proper alignment approach in broadcasting feature-maps.
Moreover, as the compression increases via down-sampling, the misalignment error of Global translation increases. Therefore, the role of Translation MOD-Alignment component in eliminating localization error becomes more significant.
It is worth mentioning that the introduced concept of feature sharing can also be utilized with a volumetric object detection scheme. In such design, the three-dimensional (3D) tensors (rather than 2D BEV image) are obtained from point clouds and are the input into the object detection component. However, the same translation MOD-Alignment procedure can be used to align the input 3D tensors prior to producing the 3D feature maps in order to enhance the performance of object detection. In 2D BEV images the Translation MOD-Alignment is performed along the two dimensions of the input image. Similarly, the alignment procedure for 3D tensors is done along three dimensions. Without loss of generality, the same formulation for alignment can be applied for the third dimension.

\subsubsection{Feature Extraction}
The feature extractor is a CNN, receiving the MOD-aligned BEV image as input and creating the feature-map of the vehicle's surrounding environment. The feature extractor architecture is shown in Table \ref{archtab} along with the details of its layers.
\subsubsection{Feature Encoder Decoder}
The structure of the feature extractor network is identical in both cooperative vehicles to project the input images of both vehicles onto an identical feature space. In this design, the feature-maps' number of channels and consequently the size of extracted feature-maps are directly dependent on the structure of the network used as feature extractor component(FEC). 
In FS-COD, the last layer of FEC is designed with respect to the communication bandwidth capacity to produce feature-maps with a specific channel size. Although maintaining the bandwidth requirement can be achieved by such design, entangling the architecture to bandwidth limitations results in inflexibility of design in adapting to different bandwidth requirements and low performance as drawbacks.

Since the parameters of the network are trained based on a specific pre-known channel size, producing feature-maps with different sizes satisfying different communication network capacity requirement is not addressed by FS-COD.
Although multiple trained FS-COD models can be deployed to share and aggregate feature-maps with different channel sizes, the practice is memory inefficient.
Additionally, reducing the number of filters at the last layer of FEC to further compress the features would result in lower object detection performance\cite{marvasti2020cooperative}. However, the feature compression in ego-vehicle would not benefit lowering the required network capacity and will only result in lower performance.
In order to make our method flexible in adapting to different bandwidth requirements and to mitigate the side effect of compression, we design a bank of CNN encoders, projecting (compressing) the transmitter vehicle feature maps onto a lower dimension.
Based on the bandwidth limitation, an encoder from the bank is selected and the produced feature-maps is fed into it.
The encoders are distinguished by the number of filters at their last layers. The number of filters at the last layer of the encoder ($C_t$ in Table \ref{archtab}) determines the size of data being shared between cooperative vehicles using our proposed approach.
The produced compressed feature-maps are transmitted along with the cooperative vehicle GPS information to other cooperative entities.
For each encoder, a corresponding decoder on receiver side projects the received compressed feature-maps to FEC feature-space.
The last layer of decoder have the same number of filters as the last layer of FEC.
This will ensure both decoder and FEC to be trained to represent identical feature spaces.
The details of training the encoder/decoder along with the FEC and detection component is described in section \ref{section:AFS-COD Training Method}.
\begin{figure*}[!t]%
\centering
\includegraphics[width=.8\textwidth,trim={50mm 10mm 30mm 10mm},clip]{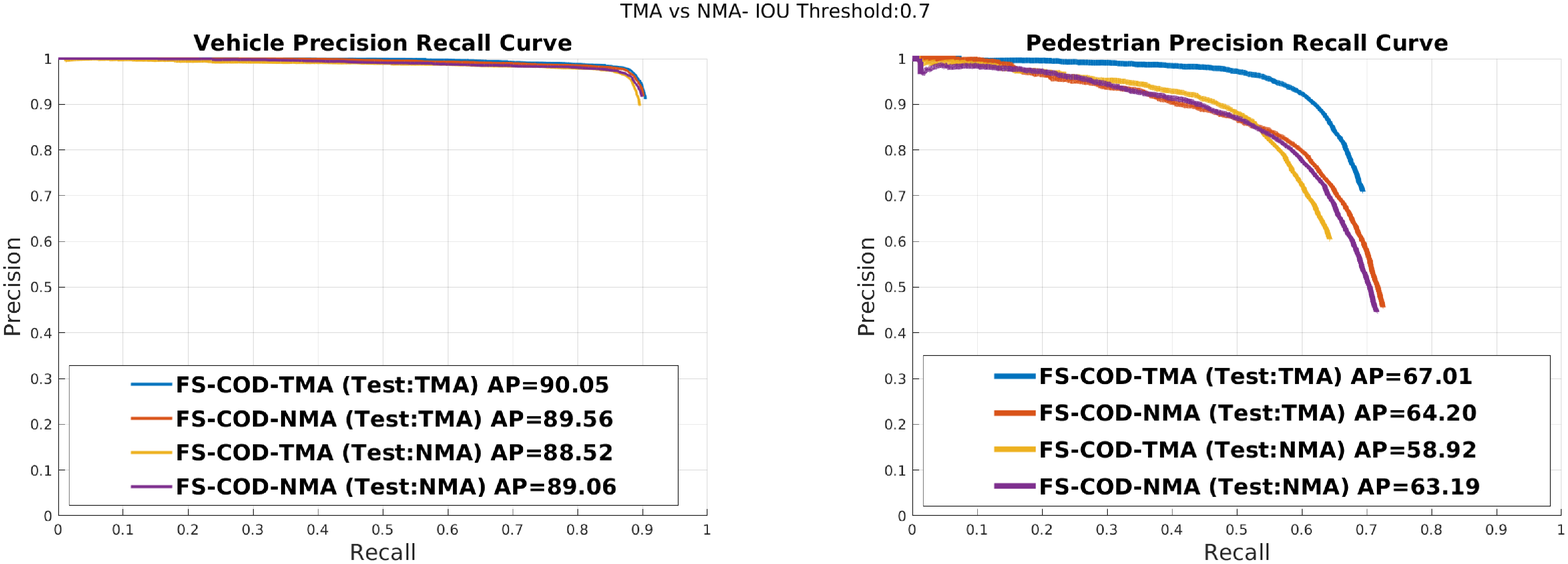}%
\caption{ (left) Precision-Recall curve of FS-COD-TMA and FS-COD-NMA in presence and absence of Translation MOD-Alignment component at test time for detection of vehicles (right) Precision-Recall curve of FS-COD-TMA and FS-COD-NMA in presence and absence of Translation MOD-Alignment component (at test time) for detection of pedestrians. 128 channels for transmission layer is chosen for all networks.}

\label{fig:TMAvsNMA}
\end{figure*}
\begin{figure*}[!t]%
\centering
\includegraphics[width=.8\textwidth,trim={50mm 10mm 30mm 10mm},clip]{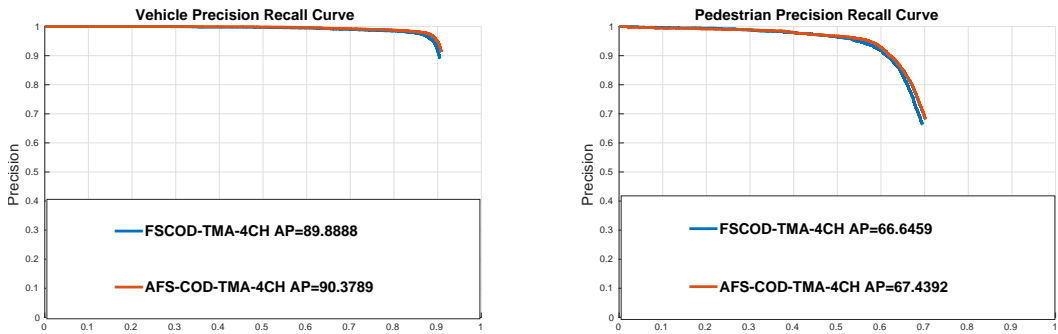}
\caption{ (left) Precision-Recall curve of FS-COD-TMA and AFS-COD for detection of vehicles (right) Precision-Recall curve of FS-COD-TMA and AFS-COD for detection of pedestrians}
\label{fig:FSCODvsVFSCOD}
\end{figure*}
\subsubsection{Global Translation Alignment}
Before accumulating the feature-maps, the decoded cooperative-vehicle's feature-map must be aligned with respect to the ego-vehicle local coordinate system.
The second phase of global alignment is a 2D image translation transformation.
\subsubsection{Element-wise summation and Object Detection}
After decoding and globally aligning the received feature-map, ego-vehicle's and aligned cooperative-vehicle's feature-maps are accumulated by an element-wise summation function.
The final stage is to detect the targets in the environment by feeding the accumulated feature-map to the object detection CNN module.
The architecture of object detection CNN is provided in Table \ref{archtab}.


\subsection{AFS-COD Training Method}
\label{section:AFS-COD Training Method}
Since AFS-COD framework consists of various encoder/decoder functions, the parameters of FEC and object detection modules should be trained to be compatible with the CNN structure and corresponding parameters of all encoder/decoders in the bank.
To satisfy the compatibility of these modules, we train all the components concurrently.
In training procedure, the FEC components of both cooperative vehicles share mutual parameters.
Additionally at each batch we randomly select an encoder/decoder function from the encoder/decoder bank to ensure compatibility of FEC and object detection modules with all members of the encoder/decoder bank.
The gradients are calculated with respect to observations of both ego and cooperative vehicles.
Assuming $g$ to be the feature accumulation function, $f$ to be the feature extractor function and $h$ the encoder-decoder function, $g$ is defined as:
\begin{equation}
    g(f(Z_1;\theta),f(Z_2;\theta)) = f(Z_1;\theta) + h(f(Z_2;\theta);\eta),
\end{equation}
Where $Z_1$ and $Z_2$ are cooperative vehicles observations, $\theta$ is the feature extractor component parameters and $\eta$ is the encoder-decoder component parameters.

\section{Experiments and Results}
\label{Section:Experiments and Results}
In this section, we provide a brief description of the dataset on which we tested our experiments. subsequently, the baseline methods are introduced. The section is concluded with evaluating the results of AFS-COD framework and comparing it to baseline methods. 

\subsection{Data Gathering and Simulation}
\label{Data Gathering and Simulation}
In order to evaluate our proposed framework, similar to our previous work, we have utilized our cooperative dataset collection tool called Volony \cite{volony2020}. This tool which works based on CARLA simulator \cite{dosovitskiy2017carla} can be used to generate simultaneous observations from the same scene in a cooperative setting. 
In our experiments, we deployed 100 vehicles and 100 pedestrians in an urban area in the simulator.
We equipped 90 vehicles with Lidar devices capturing synchronized observations from the environment. The vehicles are equipped with a 40m range LIDAR device in the simulator and the size of the 2D image resulting from BEV projection is fixed at 832$\times$832 pixels. Therefore, the image resolution is 10.4 pixels per meter(ppm).
The selected dataset includes only frames in which at least two cooperative vehicles have measurements from one or more targets in the scene. Pedestrians and vehicles in the scene are considered as targets to be detected.
The dataset consists of 1000 sample frames for each training and validation sample sets. From these sample frames, 75000 of LIDAR samples are selected for training and 10000 samples are selected for validation.
To attain fair comparison, the same set of samples was used for training and validation of AFS-COD and FS-COD schemes.

\subsection{Baseline: FS-COD with and without Translation MOD-Alignment }
\label{Baseline: Single Vehicle Object Detection}

In our previous work, it was shown that our proposed cooperative object detection method (FS-COD) outperforms single-vehicle object detection scheme. Therefore, in this work, we do not compare AFS-COD with conventional single-vehicle object detection methods. Hence, feature sharing cooperative object detection methods are considered as baselines to evaluate the performance of AFS-COD.
 
The first baseline is FS-COD without Translation MOD-Alignment, proposed in our previous work \cite{marvasti2020cooperative}, we call this FS-COD-NMA in the rest of this paper.
In order to investigate the effect of Translation MOD-Alignment in the performance of cooperative object detection, we have modified FS-COD design by adding Translation MOD-Alignment (TMA) component to its architecture. The resulted design is considered as the second cooperative object detection baseline denoted as FS-COD-TMA.

Table \ref{archtab} demonstrates the network architecture of baseline methods.

\subsection{Evaluation and Results}
\label{Evaluation and Results}

In this section, the evaluation of our proposed framework is presented. The results of AFS-COD are compared to both baseline methods discussed in the previous section. 
Average Precision, as a commonly used metric, is considered to evaluate object detection performance of the presented methods\cite{padillaCITE2020}.
In this paper, the IOU threshold of $0.7$ is selected for calculating average precision.

In order to have a fair comparison, the encoder/decoder component in AFS-COD is chosen to have an identical transmitting feature-map channel size with the selected FS-COD model. This will ensure both methods having the same bandwidth requirement.  

First, we investigate the effect of Translation MOD-Alignment in enhancing the performance of cooperative object detection via feature sharing.
For this purpose, we have trained FS-COD architecture with Translation MOD-Alignment and without Translation MOD-Alignment for creating and sharing feature-maps with 128 channels. 

In the testing procedure, we test both trained networks with and without the presence of TMA in their architecture, e.g, the trained FS-COD-TMA network is tested in presence and in absence of TMA component and are referred to as FS-COD-TMA (Test:TMA) and FS-COD-TMA (Test:NMA) respectively. 
Fig. \ref{fig:TMAvsNMA} demonstrates the effect of including Translation MOD-Alignment on precision recall curve. 
It is shown that Translation MOD-Alignment improves the performance of the networks regardless of how the network is trained. 
Additionally, discarding Translation MOD-Alignment component (at testing time) has a noticeable negative impact on the average precision of FS-COD-TMA performance.
The results decisively demonstrate the performance boost caused by Translation MOD-Alignment.
The performance improvement in detecting pedestrians is more noticeable due to the sensitivity of detection to localization error in this class of targets. 
It can be inferred that Translation MOD-Alignment gains more significance as the ratio of the size of target to area represented by fixels decreases. 
In other words, Translation MOD-Alignment becomes more beneficial as down-sampling rate increases to provide a higher compression rate.

Finally, we have investigated the effect of adding encoder and decoder on performance. As it was discussed, incorporating the encoder/decoder component will enable the framework to be able of transmitting and aggregating information with various compression rate.
We have chosen FS-COD-TMA and AFS-COD designs to transmit feature-maps with 4 channels ($C_t=4$ in Table \ref{archtab}) and compare their object detection performance. If each element of the feature-map is represented by a 32-bit single-precision floating-point and the feature-map resolution is considered to be $52\times52$, the shared data size per frame will be $40$ KB.
Fig.\ref{fig:FSCODvsVFSCOD} illustrates that AFS-COD framework slightly outperforms FS-COD-TMA in terms of average precision. Hence, it can be inferred that deploying encoder/decoder components in AFS-COD architecture not only adds flexibility in adapting to different communication bandwidth limitations but also leads to improvement in object detection performance.
Table \ref{table::bandwidth} summarizes the performance results of all the aforementioned methods.  
\begin{table}[t]
\caption{Average Precision}
\label{table::bandwidth}
\begin{center}
\begin{tabular}{ |c|c|c|c|}

\hline
 \textbf{Model}&Test&\textbf{Vehicle AP}&\textbf{Pedestrian AP} \\
\hline
FS-COD-TMA-128CH&TMA&\textbf{90.05\%}&\textbf{67.01\%}\\ \hline 
FS-COD-NMA-128CH&TMA&89.56\%&64.20\% \\ \hline 
FS-COD-TMA-128CH&NMA&88.52\%&58.92\% \\ \hline
FS-COD-NMA-128CH&NMA&89.06\%&63.19\% \\ \hline 
AFS-COD-TMA-4CH&TMA&\textbf{90.37\%}&\textbf{67.43\%} \\ \hline
FS-COD-TMA-4CH&TMA&89.88\%&66.64\% \\ \hline
\end{tabular}
\end{center}
\vspace{-5mm}
\end{table}
\section{Concluding Remarks}
We have proposed a flexible framework for cooperative LIDAR object detection, enabling the cooperative entities to share and aggregate variable sized feature-maps. The proposed framework was developed out of FS-COD by adding two novel components.

Translation MOD-Alignment method is introduced as a remedy to eliminate the inherent misalignment error of shared feature-maps. This method can be applied to any cooperative system where grids of down-sampled data are shared between participants.
The results confirm the significant role of Translation MOD-Alignment in enhancing object detection performance. Utilizing Translation MOD-Alignment method increased the performance of detecting pedestrians by $6\%$.
By deploying an encoder/decoder bank, we disentangled the architecture from communication bandwidth limitation, enabling the cooperative vehicles to efficiently encode and transmit information with respect to communication network capacity.

\label{Section:Concluding Remarks and Future work}

\bibliographystyle{IEEEtran}\small
\bibliography{main}

\end{document}